\documentclass[letterpaper, 10 pt, conference]{ieeeconf}  
\usepackage[T1]{fontenc}    

\usepackage{hyperref}       
\usepackage{url}            
\usepackage{booktabs}       
\usepackage{amsfonts}       
\usepackage{nicefrac}       
\usepackage{microtype}      
\usepackage{graphicx}
\usepackage{subfigure}
\usepackage{amsmath,amsfonts,amsthm,amssymb}
\usepackage{mathrsfs}
\usepackage{algorithm}
\usepackage{algcompatible}
\algnewcommand\algorithmicreturn{\textbf{return}}
\algnewcommand\RETURN{\State \algorithmicreturn}%

\theoremstyle{definition}
\newtheorem{definition}{Definition}

\newtheorem{theorem}{Theorem}

\newtheorem{remark}{Remark}

\DeclareMathOperator*{\argmax}{arg\,max}
\DeclareMathOperator*{\argmin}{arg\,min}

\usepackage{graphicx}
\usepackage{subfigure}
\usepackage{amsmath}
\usepackage{multirow}
\usepackage{wrapfig}

\IEEEoverridecommandlockouts                              

\title{POPO: Pessimistic Offline Policy Optimization}

\author{Qiang He$^{1,2}$, Xinwen Hou$^{1}$
	\thanks{*This work was supported by the National Natural Science Foundation of China (61379099) and the Excellent Youth Award from the National Natural Science Foundation of China  (61722312).}
	\thanks{$^{1}$Institute of Automation, Chinese Academy of Sciences, Beijing, China
	}%
	\thanks{$^{2}$School of Artificial Intelligence, University of Chinese Academy of Sciences}%
}

\begin{document}
	
	\maketitle
	\thispagestyle{empty}
	\pagestyle{empty}

\begin{abstract}
Offline reinforcement learning (RL), also known as batch RL, aims to optimize policy from a large pre-recorded dataset without interaction with the environment. This setting offers the promise of utilizing diverse, pre-collected datasets to obtain policies without costly, risky, active exploration. However, commonly used off-policy algorithms based on Q-learning or actor-critic perform poorly when learning from a static dataset. In this work, we study why off-policy RL methods fail to learn in offline setting from the value function view, and we propose a novel offline RL algorithm that we call Pessimistic Offline Policy Optimization (POPO), which learns a pessimistic value function to get a strong policy. We find that POPO performs surprisingly well and scales to tasks with high-dimensional state and action space, comparing or outperforming several state-of-the-art offline RL algorithms on benchmark tasks.
\end{abstract}

\section{INTRODUCTION}
One of the main driving factors for the success of the mainstream machine learning paradigm in open-world perception environments (such as computer vision, natural language processing) is the ability of high-capacity function approximators (such as deep neural networks) to learn inductive models from large amounts of data~\cite{resnet}~\cite{gpt-3}. Combined with deep learning, reinforcement learning (RL) has proven its great potential in a wide range of fields such as playing Atari games~\cite{dqn}, playing chess, Go and shoji~\cite{alpha}, beating human players in StarCraft~\cite{starcraft} etc. However, it turns out that reinforcement learning is difficult to extend from physical simulators to the unstructured physical real world because most RL algorithms need to actively collect data due to the nature of sequential decision making, which is very different from the typical supervised learning setting. In this paper, we study how to utilize RL to solve sequential decision-making problems from a fixed data set, i.e., offline RL, a.k.a. batch RL, which is opposite to the research paradigm of active, interactive learning with the environment. 
In the physical world, we can usually obtain static data from historical experiences more easily than dynamics data, such as scheduling a region's power supply system. There are problems with model deviation for such a scenario, and too expensive for manufacturing a simulator. Therefore, learning from static datasets is a crucial requirement for generalizing RL to a system where the data collection procedure is time-consuming, risky, or expensive.
In principle, if assumptions about the quality of the behavior policies that produced the data can be satisfied, then we can use imitation learning (IL)~\cite{imitate} to get a strong policy. However, many imitation learning algorithms are known to fail in the presence of suboptimal trajectories or to require further interaction with the environment in which the data is generated from~\cite{hester2017deep}~\cite{sun2018truncated}~\cite{chemali2015direct}. Many off-policy RL methods have proven their excellent sample-efficiency in complex control tasks or simulation environments recently~\cite{ddpg}~\cite{td3}~\cite{wd3}. Generally speaking, off-policy reinforcement learning is considered to be able to leverage any data to learn skills. However, in practice, these methods still fail when facing arbitrary off-policy data without any opportunity to interact with its environment. Even the off-policy RL method would fail given high-quality expert data produced by the same algorithm. This phenomenon goes against our intuition about off-policy RL because if shown expert data, then exploration, RL's intractable problem, no longer exists. The sensitivity of existing RL algorithms to data limits the broad application of RL. We aim to develop a completely offline RL algorithm, which can learn from large, arbitrarily static datasets. 

Our contributions are summarized as follows. Firstly, we show that a critical challenge arises when applying value-based algorithms to completely offline data that the estimation gap of the value function. When we evaluate the value function, the inability to interact with the environment makes it unable to eliminate the estimation gap through the Bellman equation. This gap can lead to the value function's catastrophic estimation issue for data that the actions do not appear in the data set. Secondly, We propose a novel offline policy optimization method, namely Pessimistic Offline Policy Optimization (POPO), where the policy utilizes a pessimistic distributional value function to approximate the true value, thus learning a strong policy. Finally, we demonstrate the effectiveness of POPO\footnote{Refer to \url{https://github.com/sweetice/POPO} for our code.} by comparing it with SOTA offline RL methods on the MuJoCo locomotion benchmarks~\cite{gym}. Furthermore, we conduct fine-grained experiments to verify POPO reflects the principles that informed its development. 
\begin{figure}[ht]
	\centering
	\includegraphics[width=3.3in]{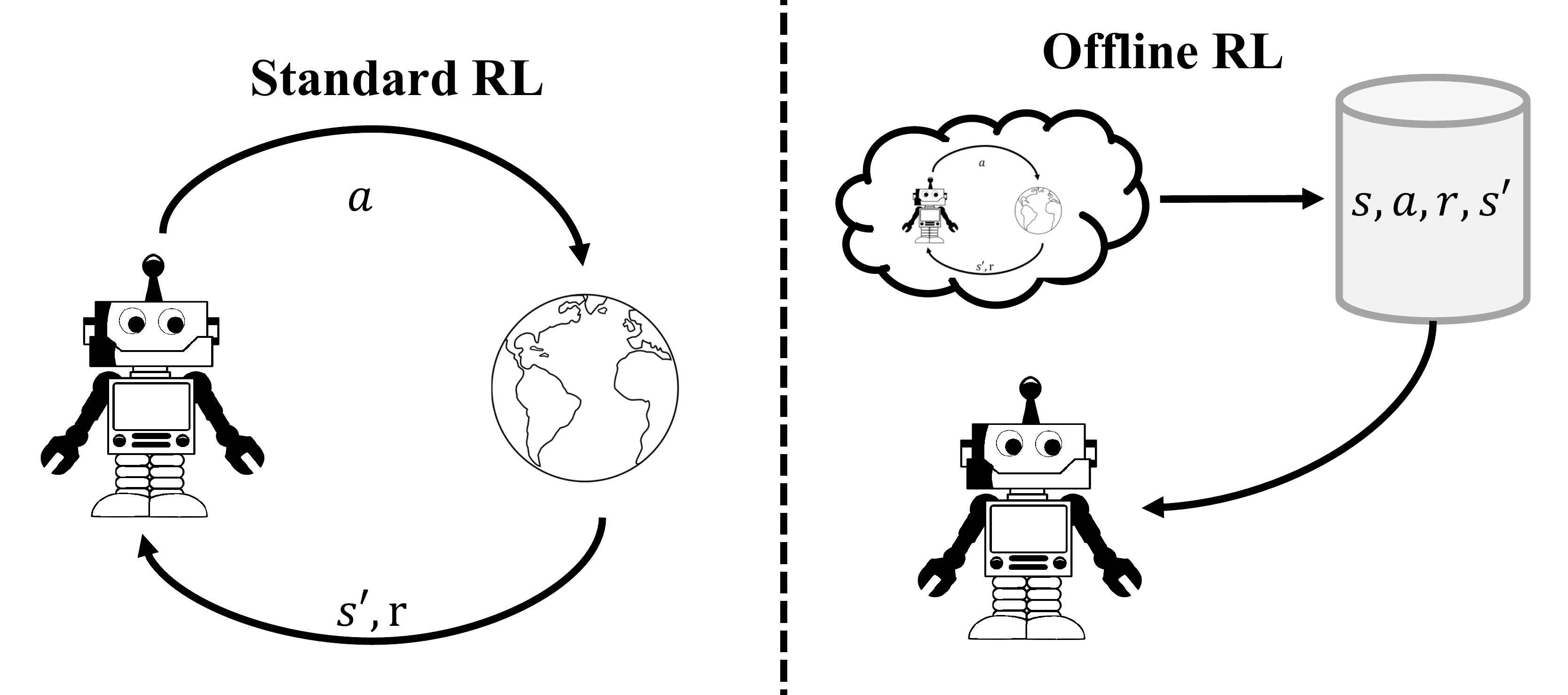}
	\caption{Difference between standard RL and offline RL. Left: Standard RL interacts with its environment directly. Right: Offline RL learns from pre-recorded dataset.}
	\label{fig:offline_and_online_RL}
\end{figure}

\section{RELATED WORK} 
Imitation learning~\cite{imitate} methods study how to learn a policy by mimicking expert experience demonstrations. IL has been combined with RL, either by learning from demonstrations~\cite{kim2013learning}~\cite{chemali2015direct}~\cite{piot2014boosted}, or using deep RL extensions~\cite{hester2017deep}~\cite{vecerik2017leveraging}, or using variants policy gradient methods~\cite{ho2016model}~\cite{sun2018truncated}. Although this family of methods has proven its efficiency, it is still insufficient in the face of fully offline datasets. They either require interaction with the environment or need high-quality data. These requirements are unable to meet under offline setting, making the use of imitation learning from offline data impractical~\cite{bcq}. How to deal with the impact of noise is also an urgent area in imitation learning~\cite{nair2018overcoming}~\cite{evans2015learning}. Gao et al.~\cite{gao2018reinforcement} introduced an algorithm that learns from imperfect data, but it is not suitable for continuous control tasks. We borrow from the idea of imitation learning and introduce a generative model into POPO, which gives our model the potential of rapid learning. 

For unlimited data, some offline RL methods have proven their convergence, such as using non-parametric function approximation methods~\cite{gordon1995stable} and kernel methods~\cite{ormoneit2002kernel}. Fitted Q-iteration, using function approximation methods, such as decision trees~\cite{ernst2005tree}, neural networks~\cite{riedmiller2005neural}, cannot guarantee convergence in the offline setting. Recently, many offline RL algorithm combined with deep learning have received significant attention ~\cite{bcq}~\cite{bear}~\cite{spibb}~\cite{rem}~\cite{brac}\cite{algaedice}~\cite{crr}~\cite{cql}~\cite{wang2018exponentially}. Offline RL suffers from the problem of the distribution shift, a.k.a. out-of-distribution (OOD) actions. Specifically, the target of the Bellman backup operator utilizes actions sampled from the learned policy in the policy evaluation and policy improvement process, which may not exist in the datasets. In the sense of batch-constrained, the BCQ algorithm~\cite{bcq} can ensure that it converges to the optimal policy under the given consistent datasets.  Bootstrapping error accumulation reduction (BEAR) algorithm \cite{bear} uses maximum mean discrepancy (MMD)~\cite{mmd} to constrain the support of learned policy close to the behavior policy. Safe policy improvement with baseline bootstrapping (SPIBB)~\cite{spibb}, similar to BEAR, constrains the support of learned policy w.r.t. behavior policy. Behavior regularized actor-critic (BRAC)~\cite{brac} is an algorithmic framework that generalizes existing approaches to solve the offline RL problem by regularizing the behavior policy. AlgaeDICE~\cite{algaedice} is an algorithm for policy gradient from arbitrary experience via DICE~\cite{dualdice}, which is based on a linear programming characterization of the Q-function. Critic Regularized Regression (CRR)~\cite{crr} algorithm can be seen as a form of filtered behavior cloning where data is selected based on the policy's value function. Conservative Q-learning (CQL)~\cite{cql} aims to learn a conservative Q-function such that the expected value of a policy under this Q-function lower-bounds its true value. Random ensemble mixture (REM)~\cite{rem} uses random convex combinations of value functions to learn a static data set. Besides, REM proves that distributional RL can learn a better policy than the conventional form in offline setting. But there is a controversy that the success comes from the massive amount of data resulting in actions induced by concurrent policy always in the datasets~\cite{bcqbechmarking}. However, their work did not consider distributional value functions to control the attitude towards the OOD actions, thus improving policy. We recommend that readers check IQN algorithm~\cite{iqn}. These methods focus on how to deal with OOD actions.
\section{BACKGROUND}
\begin{figure*}[ht]
	\centering
	\hspace{-0.3in}
	\subfigure{
		\includegraphics[width=1.7in]{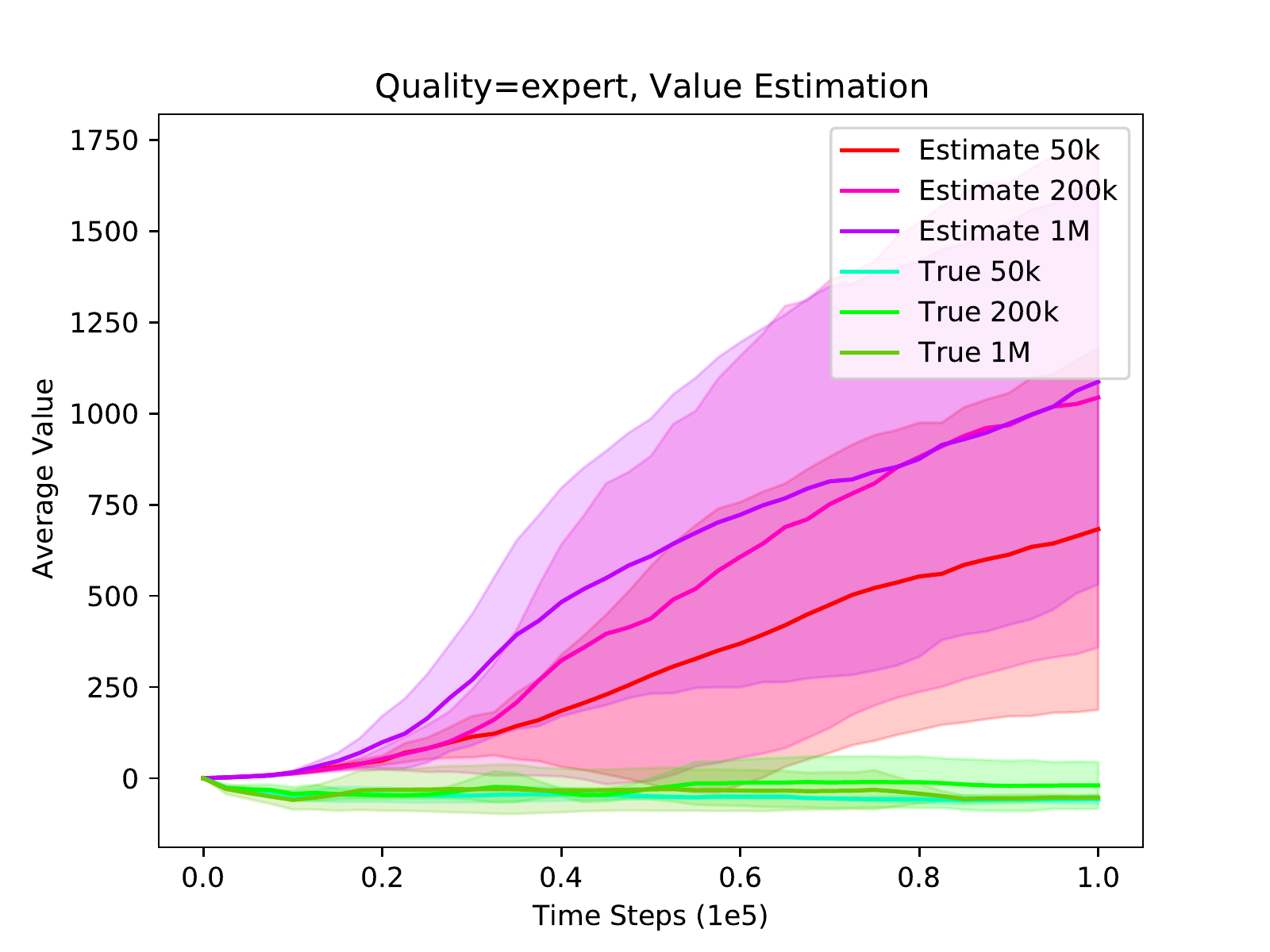}
	}
	\hspace{-0.3in}
	\subfigure{
		\includegraphics[width=1.7in]{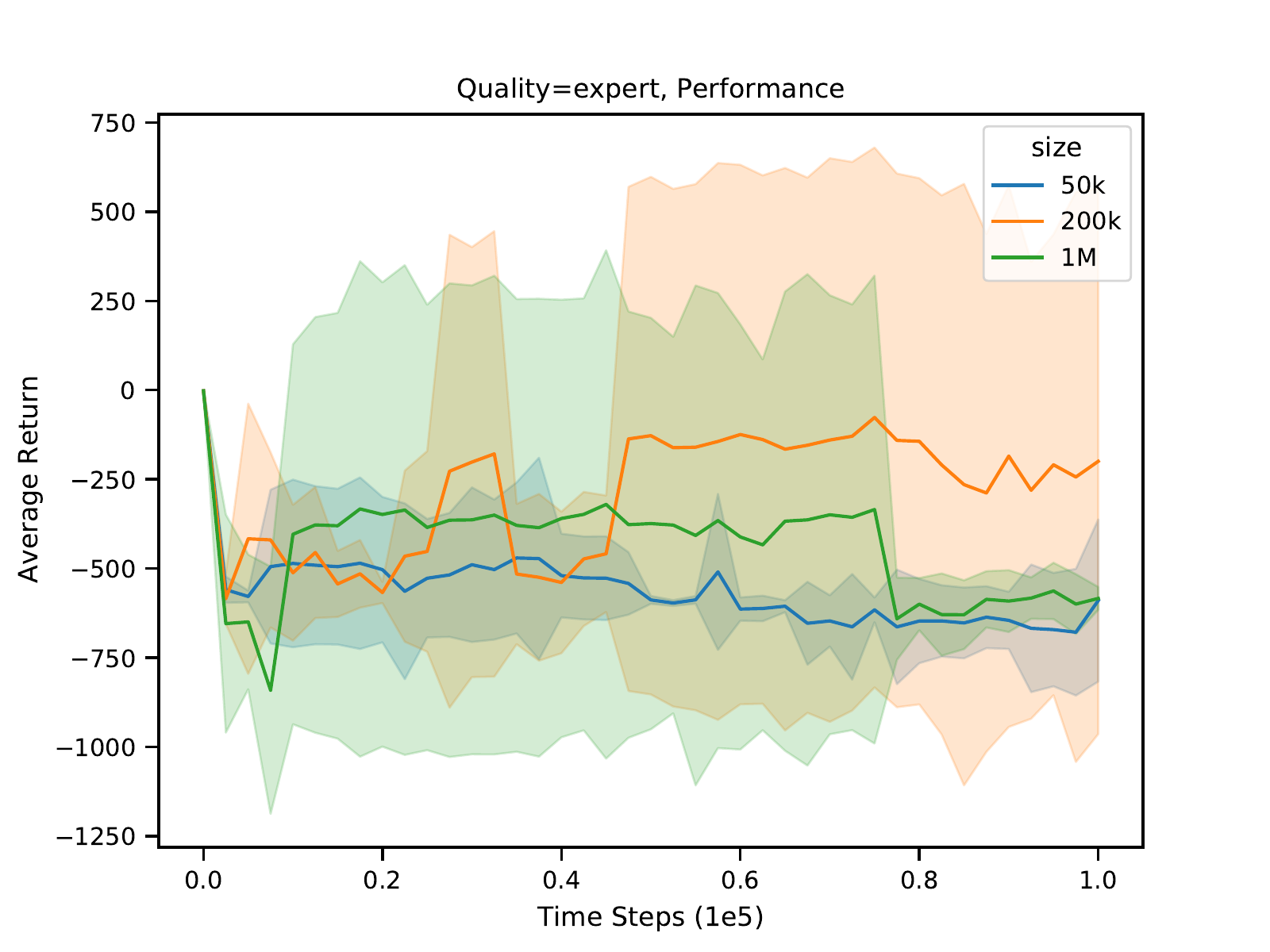}
	}
	\hspace{-0.3in}
	\subfigure{
		\includegraphics[width=1.7in]{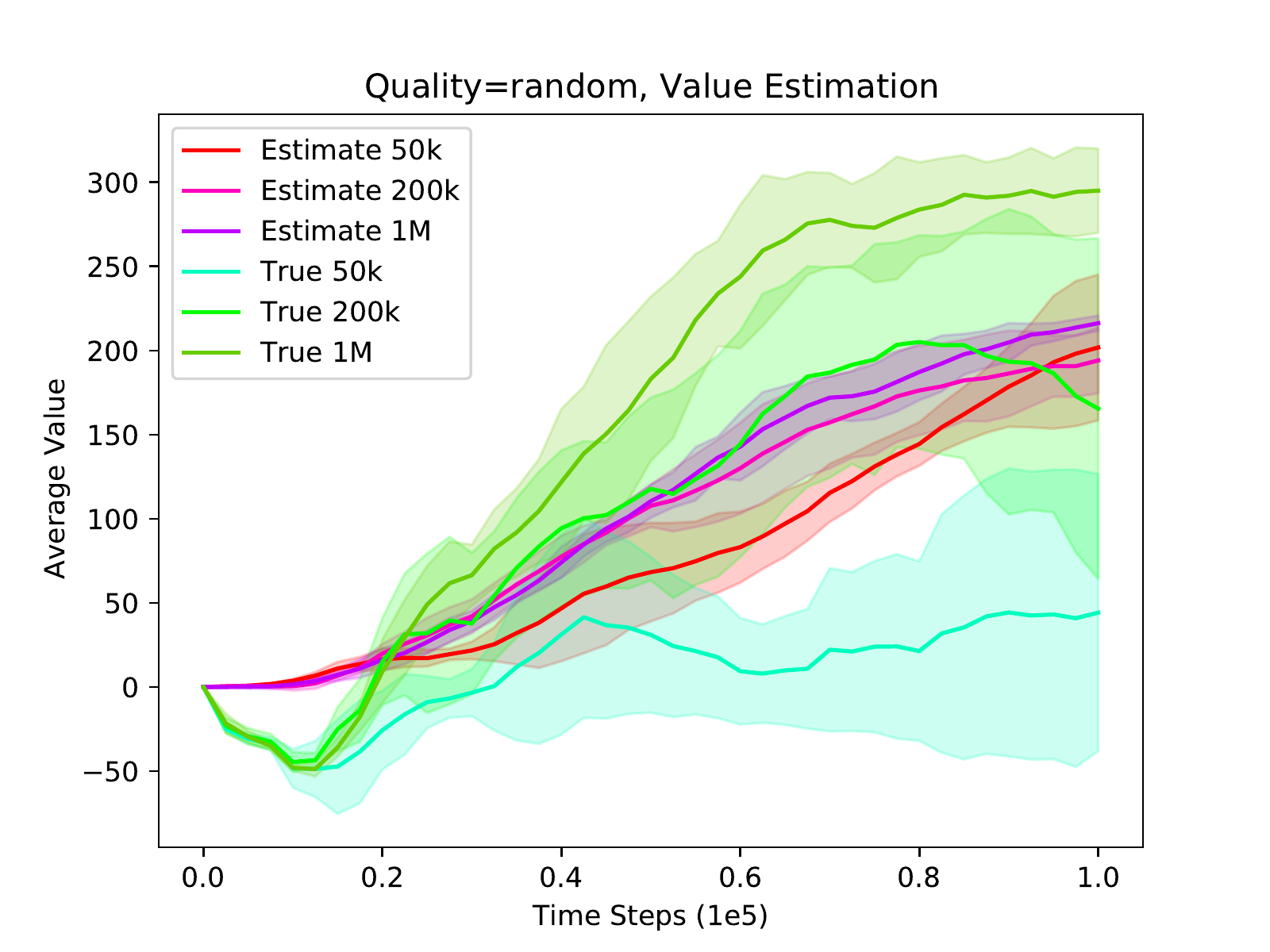}
	}
	\hspace{-0.3in}
	\subfigure{
		\includegraphics[width=1.7in]{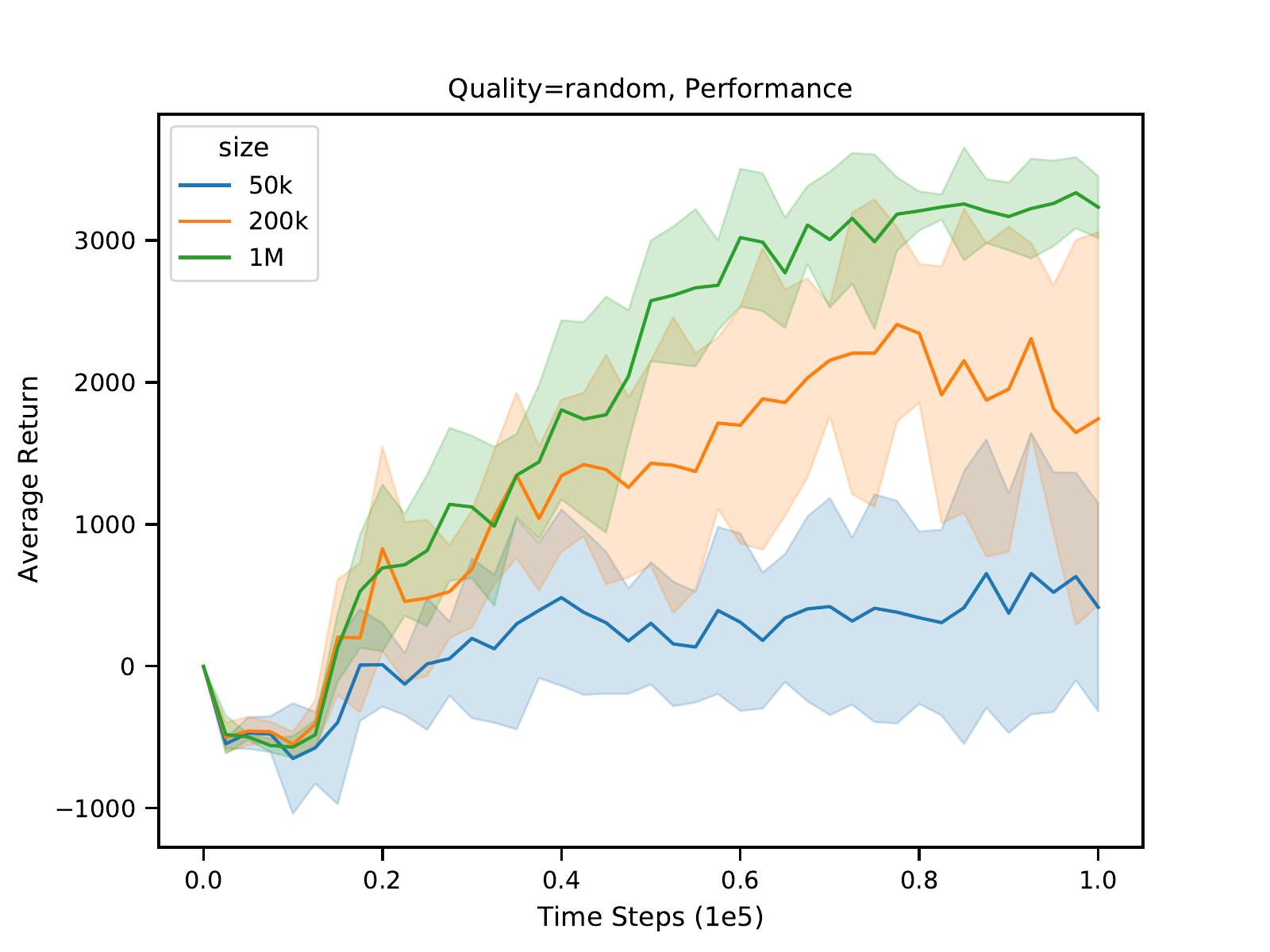}
	}
	
	\caption{The relationship between data quality, quantity, and corresponding average return. We train TD3 algorithm on MuJoCo halfcheetah-v2 environment over five random seeds. 'Estimate 50k' means the curve shows the agent's value estimation on the size=50 dataset. To maximally control the influence of random factors, we control the random seed of each experiment. The shaded area represents a standard deviation.}
	\label{fig:various_quality_env_type}
\end{figure*}
The optimization goal of RL is to get an optimal policy by interacting with its environment in discrete timesteps. We formalize the standard RL paradigm as a Markov Decision Process (MDP), defined by a tuple $(\mathcal{S ,A ,R,} p,\rho_0 ,\gamma)$ with state space $\mathcal{S}$, action space $\mathcal{A}$, reward function $\mathcal{R: S \times A \times S}\rightarrow \mathbb{R}$, transition probability function $p(s',r|s,a)$, initial state distribution $\rho_0$, and discount factor $\gamma \in [0,1)$. At each time step $\mathit{t}$, the agent receives a state $\mathit{s} \in \mathcal{S}$ and selects an action $\mathit{a} \in \mathcal{A}$ with respect to its policy  $\pi : \mathcal{S} \rightarrow \mathcal{A}$, then receiving a reward signal $\mathit{r}$ and a new state $\mathit{s'}$ from its environment. A four-element tuple $(s,a,r,s')$ is named transition. The optimization problem in RL is to maximize the cumulative discounted reward, defined as $R_t = \sum_{i=t}\gamma^{i-t}r(s_i, a_i)$ with $\gamma$ determining the priority of recent rewards. Here the return depends on the actions, thus on the policy $\pi$, deterministic or stochastic. Typically, the action-value function, a.k.a. Q-function, critic, is defined as $Q(s,a) = \mathbb{E}_{p, \pi} [R_t | s_0=s, a_0=a]$ which measures the quality of an action $a$ given a state $s$. State-value function, a.k.a value function, $V$-function, is defined as $V(s) = \mathbb{E}_{p,\pi} [R_t|s_0=s]$ measuring the quality of an individual state $s$. Both $Q$-function and $V$-function can be applied to evaluate the policy and further guide the agent to learn a higher quality value, i.e., a better policy. For a given policy $\pi$, the $Q$-function can be estimated recursively by Bellman backup operator~\cite{rl}:
\begin{equation}
Q^{\pi}(s,a) = r + \gamma \mathbb{E}_{s',a'} [Q^{\pi}(s',a')],
\label{eq:Q-learning}
\end{equation}
where $a'\sim \pi(s')$. The Bellman operator is gamma-contraction when  $\gamma \in [0,1)$ with a unique fixed point $Q^{\pi}(s,a)$. We can recover the optimal policy through the corresponding optimal value function $Q^*(s,a)=\max_{\pi}Q^{\pi}(s,a)$ in discrete action space. 
When we apply RL to large state space or continuous state space, the value function can be approximated by neural networks, which is called Deep $Q$-networks~\cite{dqn}. The $Q$-function is updated by $r+\gamma Q(s', \pi(s');\theta')$, where $\pi(s')=\arg\max_{a'} Q(s',a';\theta')$ and $\theta'$ is a delayed copy of $\theta$. Generally, we sample mini-batch transitions from replay buffer $\mathcal{B}$~\cite{replay_buffer} and feed the data into the deep $Q$-networks. In offline policy optimization setting, we consider the buffer $\mathcal{B}$ static and no further data can be added into itself and call it datasets $\mathcal{D}$. A vitally significant improvement of DQN is soft update. When updating the network, we freeze a target network $Q(\cdots;\theta')$ to stabilize the learning processing further. The frozen network are updated by $\theta' = \eta \theta + (1-\eta) \theta'$ every specific time steps $t$, where the $\eta$ is a small scalar~\cite{dqn}.
In continuous action space, the $\arg \max$ operator in Equation~\ref{eq:Q-learning} is intractable. Thus Sutton et al. \cite{policy_gradient} introduced policy gradient method. Combined with aforementioned value-based method, actor-critic method was introduced, which is widely used in the field of deep RL. When we train an actor-critic agent, the action selection is performed through a policy network $\pi(\cdot;\phi)$, a.k.a. actor, and updated w.r.t. a value network~\cite{rl}. Silver et al.~\cite{dpg} proposed deterministic policy gradient theorem to optimize policy:
\begin{equation}
\phi \leftarrow \arg \max_{\phi} \mathbb{E}_{s} \left[Q^{\pi}(s,\pi(s;\phi);\theta)\right],
\label{eq:DDPG_Value_function_update}
\end{equation}
which corresponds to optimizing the $Q$-function $Q^{\pi}(\cdot, \cdot;\theta)$ by the chain rule. When combined with tricks in DQN~\cite{dqn}, this algorithm is referred to as deep deterministic policy gradients (DDPG)~\cite{ddpg}. 

\section{DIAGNOSING VALUE FUNCTION ESTIMATION}
Offline reinforcement learning suffers from OOD actions. Specifically, the target of the Bellman backup operator utilizes actions generated by the learned policy in the policy evaluation and policy improvement process. However, the generated actions may not exist in the dataset. Thus, we cannot eliminate the error through the Bellman update. Both the value-based method and policy gradient methods would fail for this reason. Hasselt et al.~\cite{ddqn} observed that overestimation occurs in the DQN algorithm. We argue that the analogous phenomenon also occurs in offline scenario but for the different underlying mechanism. These two phenomena are coupled with each other, making the value function more difficult to learn than online setting. In the standard reinforcement learning setting, these errors due to erroneous estimation could be eliminated through the agent's exploration to obtain a true action value and then updated by the Bellman backup operator. But for offline setting, this error cannot be eliminated due to the inability to interact with the environment. Furthermore, due to the backup nature of the Bellman operator, the error would gradually accumulate, which would eventually cause the value function error to become larger, leading to the failure of policy learning. Some algorithms train policies through optimizing the Q-value indirectly. And some actor-critic style methods optimize the policies directly but are assisted by the value function. Therefore, out-of-distribution actions harm these RL algorithms' performance in offline setting. We call the aforementioned error estimation gap. 
\begin{definition}
	We define estimation gap for policy $\pi$ in state $s$ as $\delta_{\text{MDP}}(s) = V^\pi(s)-V^\pi_{\mathcal{D}}(s)$ where $V^\pi(s)$ is true value and $V^\pi_{\mathcal{D}}(s)$ is estimated on dataset $\mathcal{D}$.
\end{definition}
\begin{theorem}
	Given any policy $\pi$ and state $s$, the estimation gap $\delta_{\text{MDP}}(s)$ satisfies the following Bellman-like equation:
	\centering
	\begin{equation*}
	\begin{aligned}
	\delta_{\text{MDP}} (s)=& \sum_{a}\pi(a|s) \sum_{s',r}\left[p(s',r|s,a)-p_{\mathcal{D}}(s',r|s,a)\right] \big(r+\\\gamma V^{\pi}_{\mathcal{D}}(s')\big) 
	& + \gamma\sum_{a}\pi(a|s)\sum_{s',r}p(s',r|s,a) \delta_{\text{MDP}}(s')
	\end{aligned}
	\end{equation*} 
\end{theorem}
\begin{proof}
	We can prove it by expanding this equation through the definition of the $V$ function.
		\begin{equation*}
	\begin{aligned}
	\delta_{\text{MDP}} (s) = & V^{\pi}(s) - V^{\pi}_{\mathcal{D}}(s) \\
	= &\mathbb{E}[r(s,a,s') + \gamma V^{\pi}(s')] - V^{\pi}_{\mathcal{D}}(s) \\
	= \sum_{a}\pi(a|s) &\sum_{s',r}p(s',r|s,a)[r+\gamma V^{\pi}(s')] - V^{\pi}_{\mathcal{D}}(s) \\
	=  \sum_{a}\pi(a|s) &\sum_{s',r}p(s',r|s,a)[r+\gamma (V^{\pi}_{\mathcal{D}}(s')+\delta_{\text{MDP}}(s'))] \\
	 - \sum_{a}\pi(a|s) &\sum_{s',r}p_{\mathcal{D}}(s',r|s,a)[r+\gamma V^{\pi}_{\mathcal{D}}(s')] 	\\
	=  \sum_{a}\pi(a|s) &\sum_{s',r}\left[p(s',r|s,a)-p_{\mathcal{D}}(s',r|s,a)\right]r \\ 
	+ \sum_{a}\pi(a|s) &\sum_{s',r}p(s',r|s,a)\gamma \delta_{\text{MDP}}(s')  \\
	 + \gamma \sum_{a}\pi(a|s) &\sum_{s',r}\left[p(s',r|s,a)-p_{\mathcal{D}}(s',r|s,a)\right]V^{\pi}_{\mathcal{D}}(s') \\
	= \sum_{a}\pi(a|s) &\sum_{s',r}\left[p(s',r|s,a)-p_{\mathcal{D}}(s',r|s,a)\right] \big(r+\\\gamma V^{\pi}_{\mathcal{D}}(s')\big) 
	& + \gamma\sum_{a}\pi(a|s)\sum_{s',r}p(s',r|s,a) \delta_{\text{MDP}}(s')
	\end{aligned}
	\end{equation*}
	where the transition probability function of data set $\mathcal{D}$ is defined as $p_{\mathcal{D}}(s',r|s,a)=\frac{N(s,a,s',r)}{\sum_{s',r}N(s,a,s',r)}$, $N(s,a,s',r)$ is the number of the transition observed in data set $\mathcal{D}$.
\end{proof}
Theorem 1 shows that the estimation gap is a divergence function w.r.t. the transition distributions, which means if the policy carefully chooses actions, the gap can be minimized by visiting regions where the transition probability is similar. 
\begin{remark}
	For any reward function, $\delta_{\text{MDP}}=0$ if and only if $p(s',r|s,a)=p_{\mathcal{D}}(s',r|s,a).$
\end{remark}
\subsection{Does this phenomenon occurs in practical?} 
\begin{figure}[ht]
	\centering
	\subfigure{
		\includegraphics[width=1.7in]{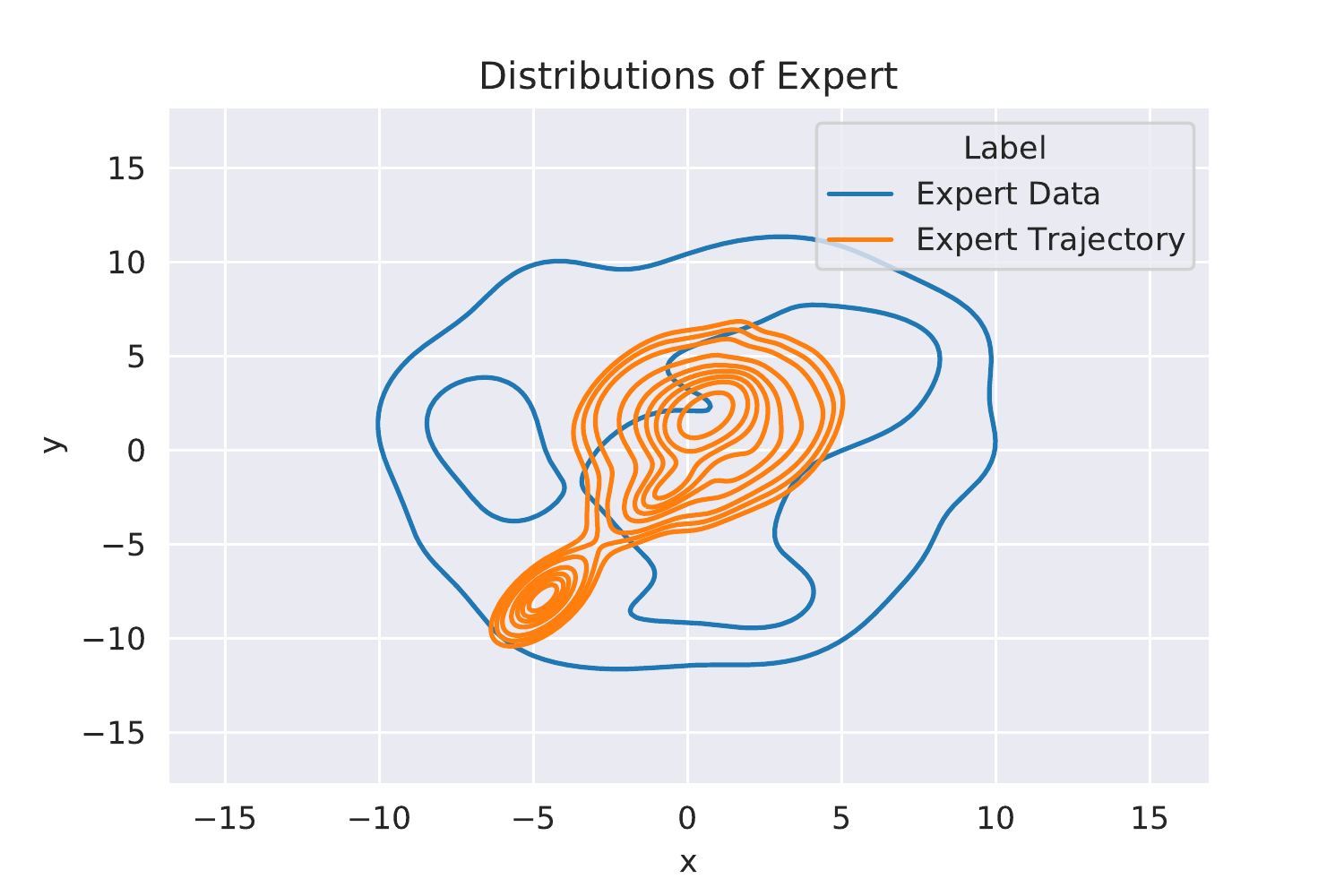}
	}
	\hspace{-0.3in}
	\subfigure{
		\includegraphics[width=1.7in]{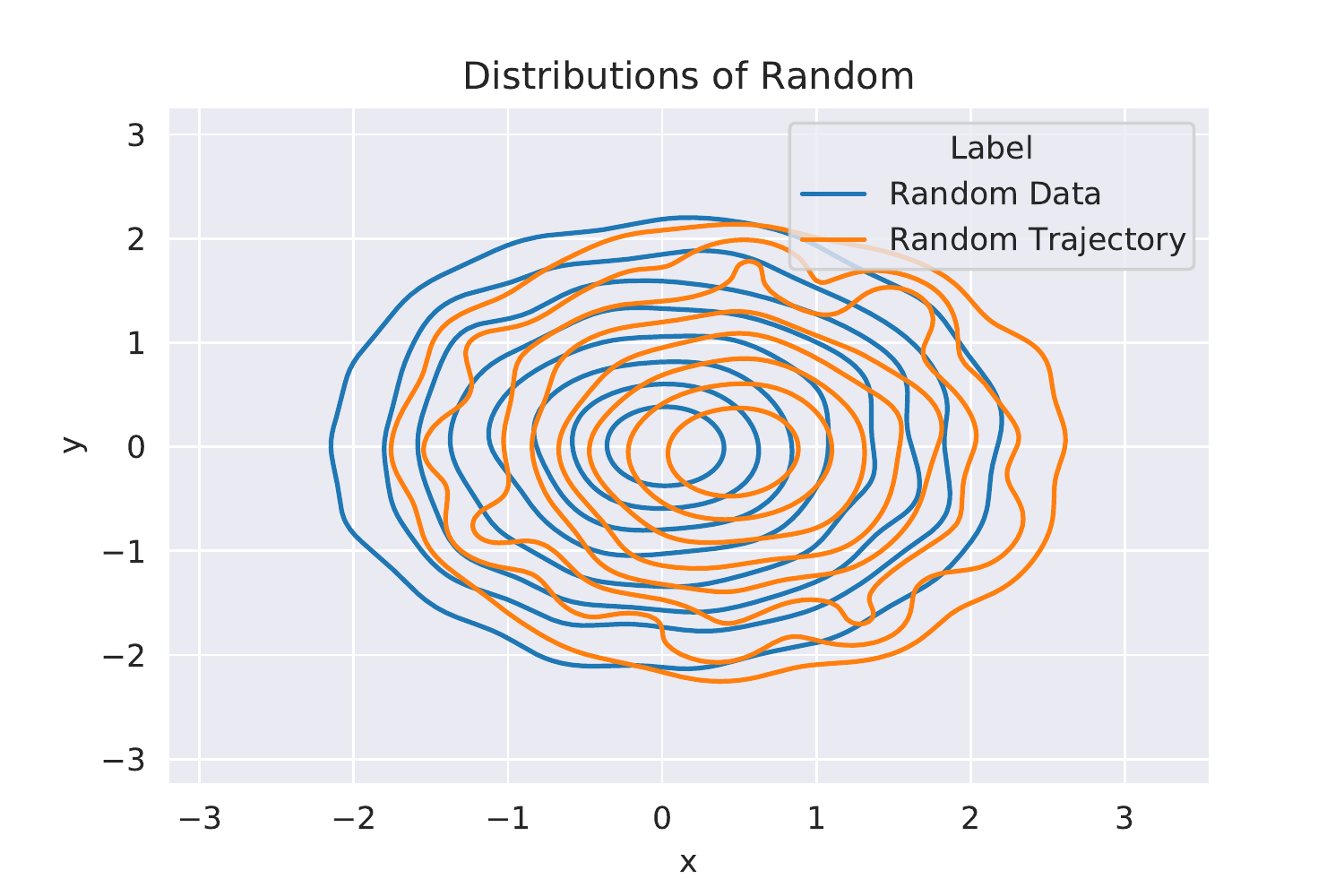}
	}
	\caption{Visualization of data generated by the halfcheetah-v2 environment. Left: expert data, visiting the different area. Right: random data, visiting a similar area. The state space is 17-dim, and the action space is 6-dim. We concat a trajectory as a vector, and we reduce the trajectories with a dimension of 23k to a two-dimensional plane.}
	\label{fig:visualize_datasets}
\end{figure}

We utilize datasets~\cite{d4rl} of different qualities and sizes to verify our analysis. We train TD3, a SOTA algorithm on continuous control tasks, on halfcheetah-v2 environment in offline setting. We show results in Figure~\ref{fig:various_quality_env_type}. Surprisingly, training on random data gives us a better average return than expert data. Observing its value function, we find that for expert data, as the data set capacity increases, the estimated value function deviates more and more from the true value, which verifies there does exist an estimation gap. The erroneous estimation of the value function further leads to the failure of policy learning. Why can random data learn better? The theory above inspires us that if the policy chooses actions carefully, it can eliminate the estimation gap by visiting regions with similar transition probability, suggesting that the phenomenon may be due to the large difference in the visited state. Thus, we visualizing~\cite{tsne} the distributions of the datasets and trajectories in Figure~\ref{fig:visualize_datasets}. We collect five trajectories every 5,000 training steps. Expert/random trajectory means we train TD3 on expert/random dataset in offline setting. We find that the TD3 agent does visit a similar area on random data. Still, for expert data, the agent visits different area from expert data even though they have the same origin, which is consistent with our theory.
\section{PESSIMISTIC OFFLINE POLICY OPTIMIZATION}
\begin{figure}[ht]
	\centering
	\includegraphics[width=3.4in]{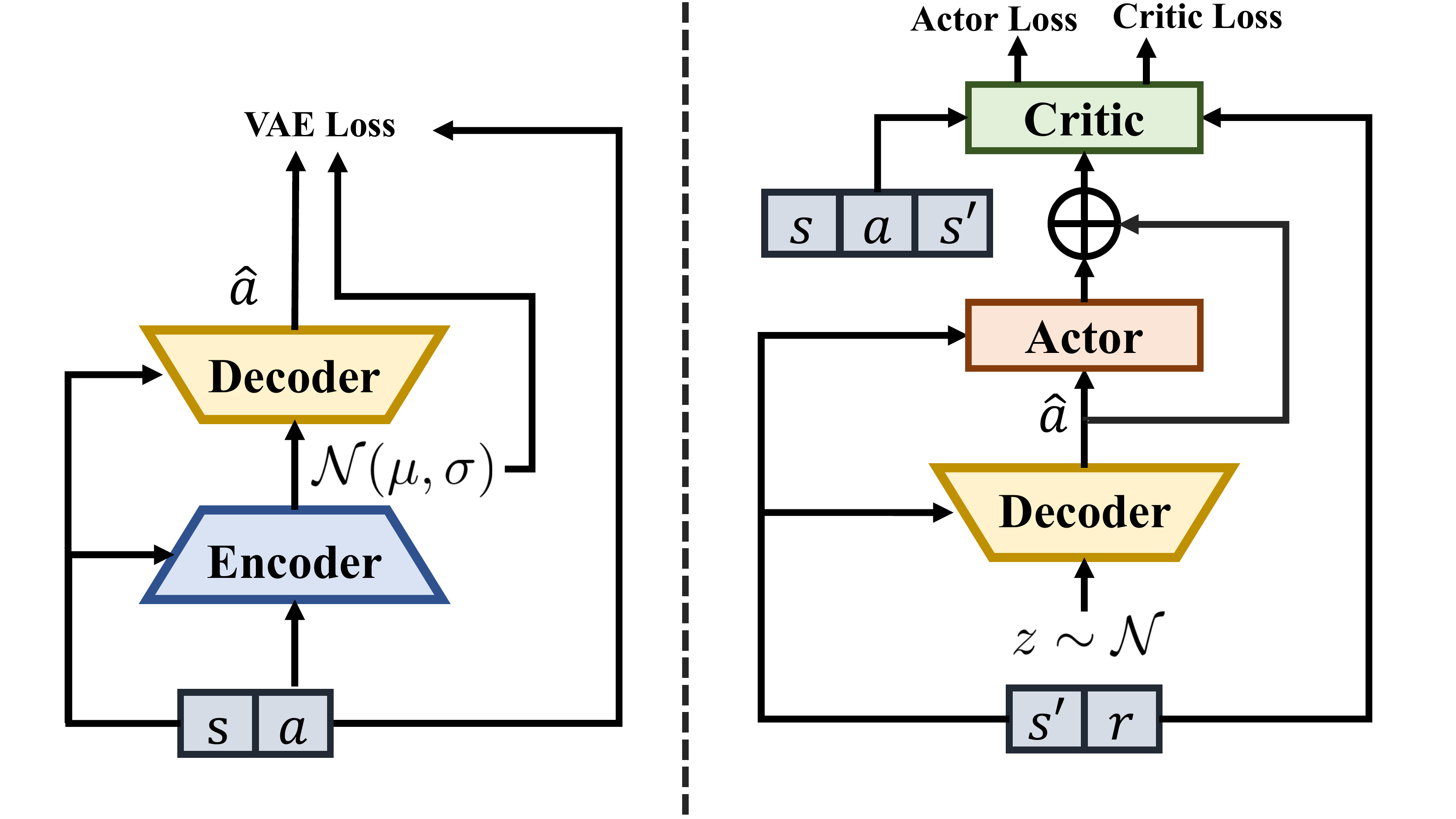}
	\caption{The architecture of POPO. Left: Conditional VAE Optimization. Right: Policy Optimization.}
	\label{fig:dataflow}
\end{figure}
Our insight is if the agent could maintain a pessimistic attitude towards the actions out of the support of behavior policy when learning the value function, then we can suppress the estimation gap of the value function outside the data set so that the algorithm can obtain a more melancholic value function to learn a strong policy through an actor-critic style algorithm. To capture more information on the value function, we utilize distributional value function~\cite{c51}~\cite{iqn}, which has proved its superiority in the online learning setting. 
\subsection{Pessimistic Value Function}
\begin{figure}[h]
	\centering
	\subfigure{
		\includegraphics[width=1.7in]{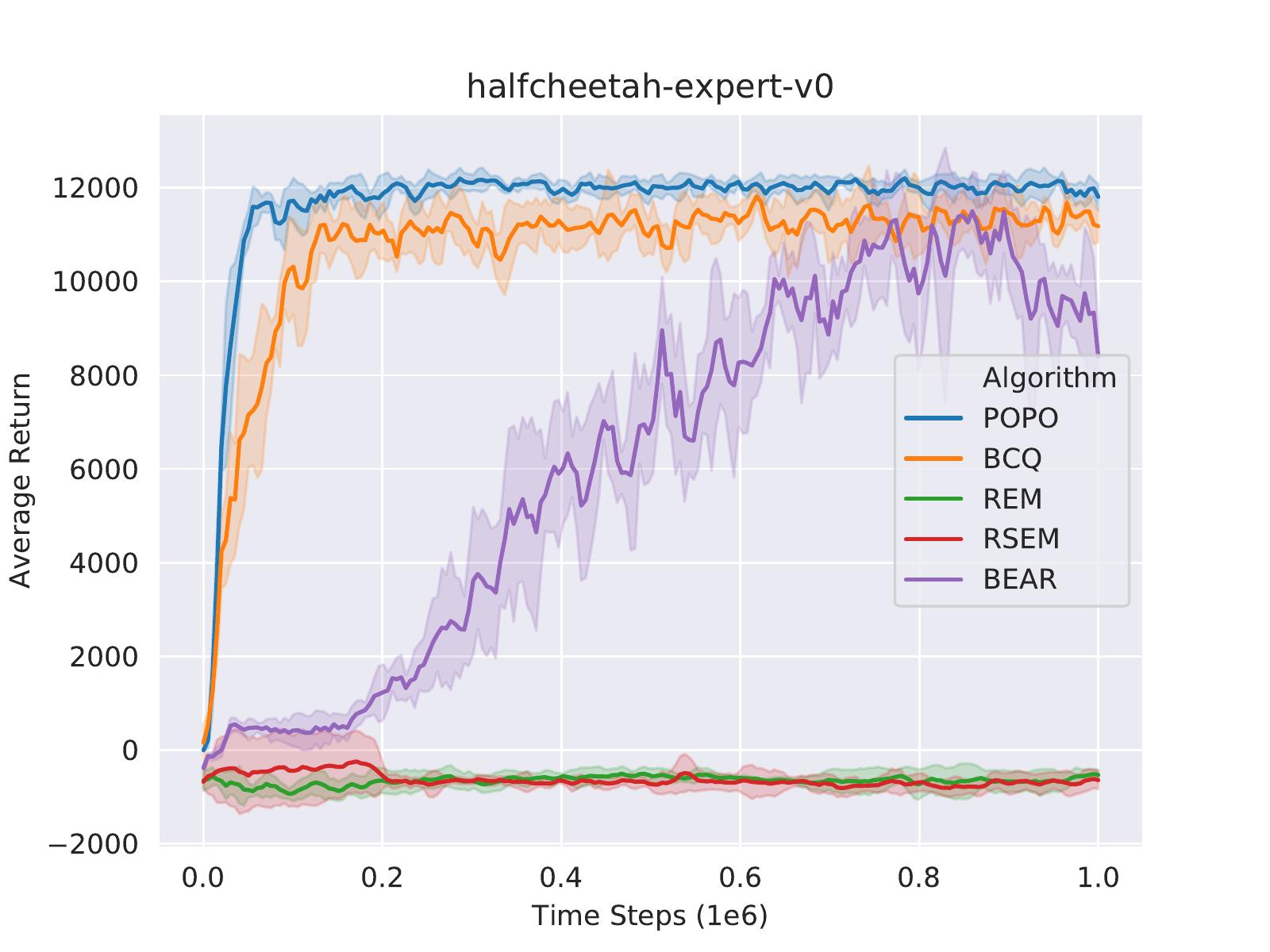}
	}
	\hspace{-0.3in}
	\subfigure{
	\includegraphics[width=1.7in]{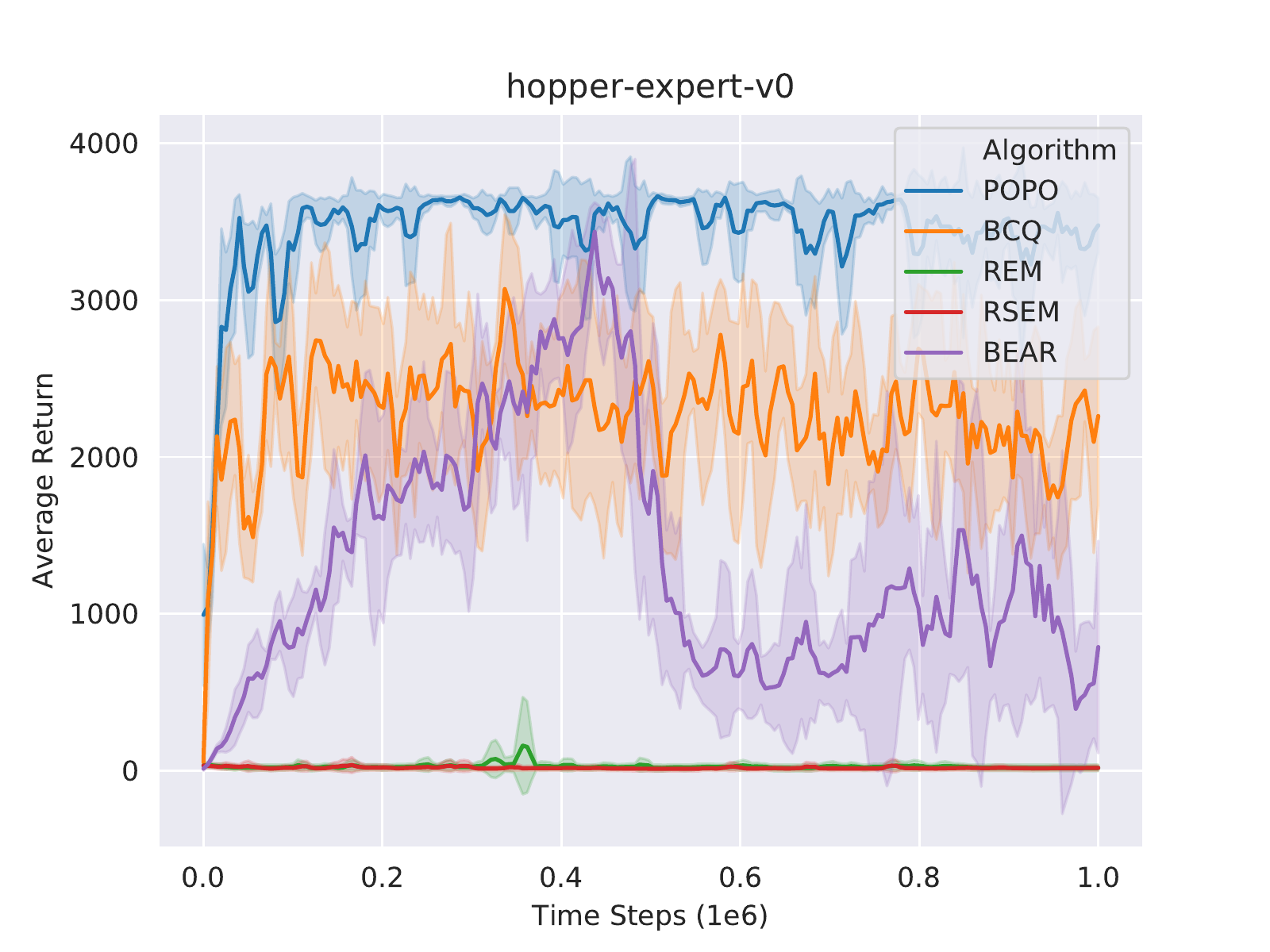}
	}  
	\subfigure{
		\includegraphics[width=1.7in]{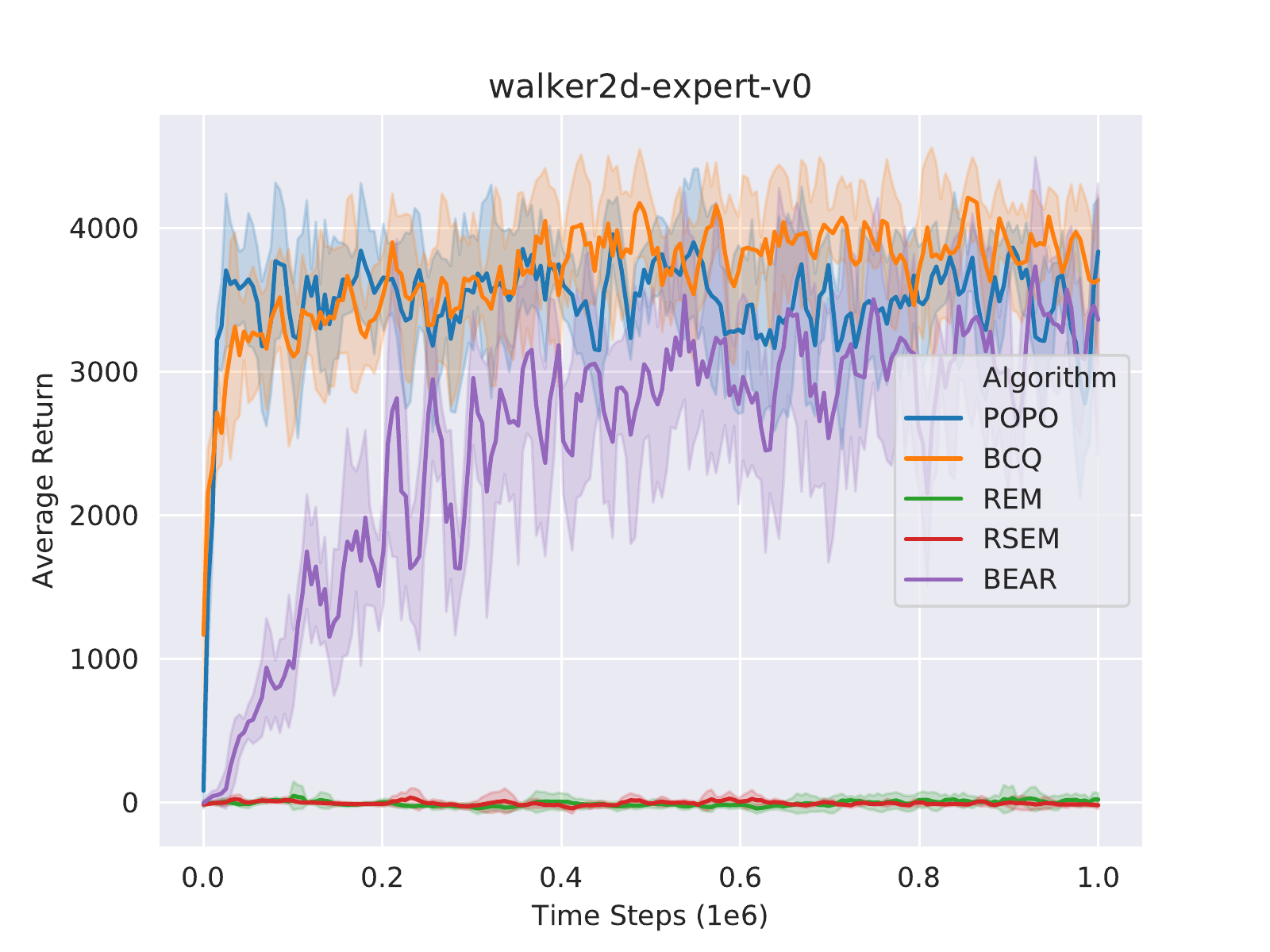}
	}
	\hspace{-0.3in}
	\subfigure{
		\includegraphics[width=1.7in]{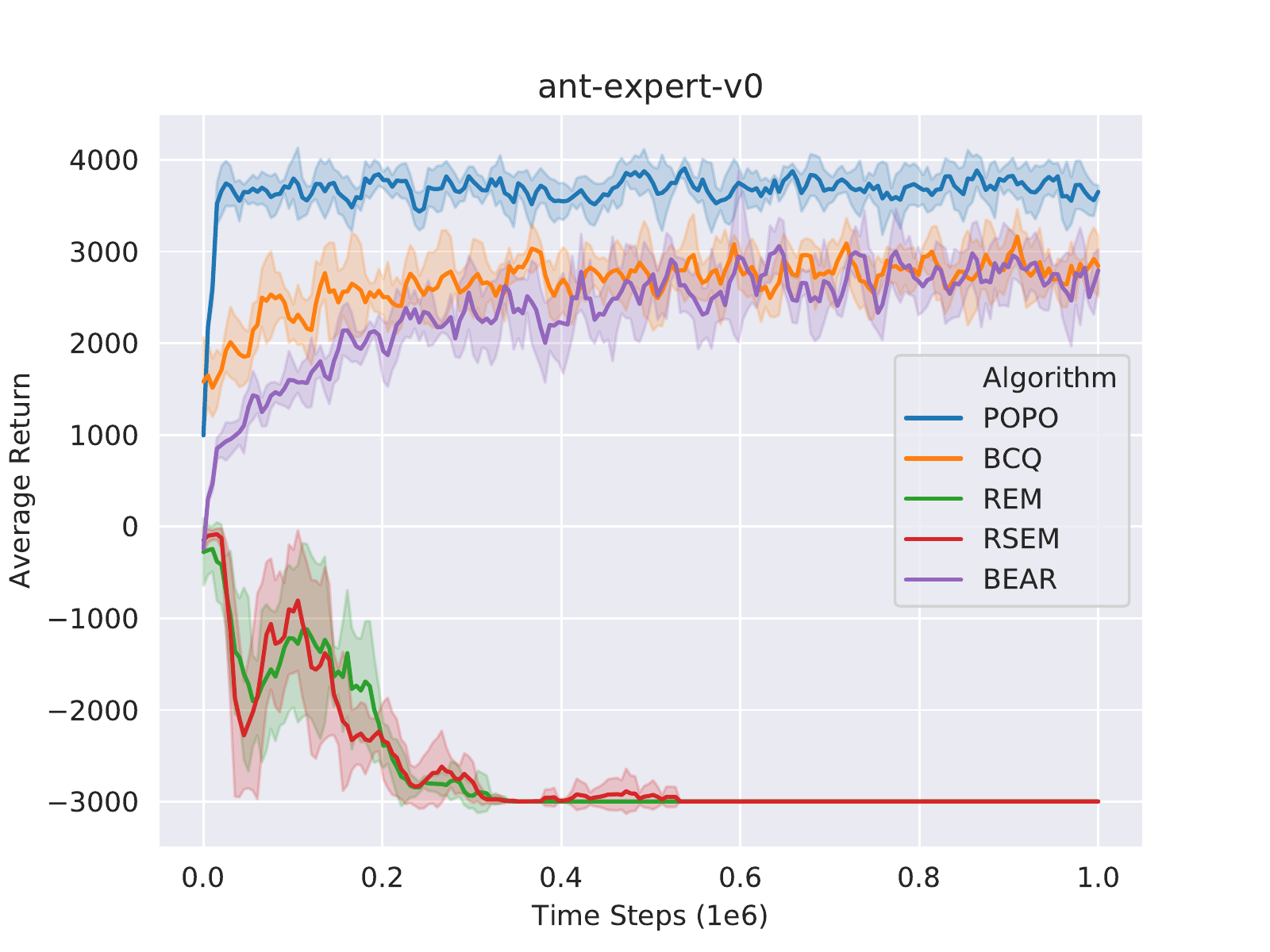}
	}

	\subfigure{
		\includegraphics[width=1.7in]{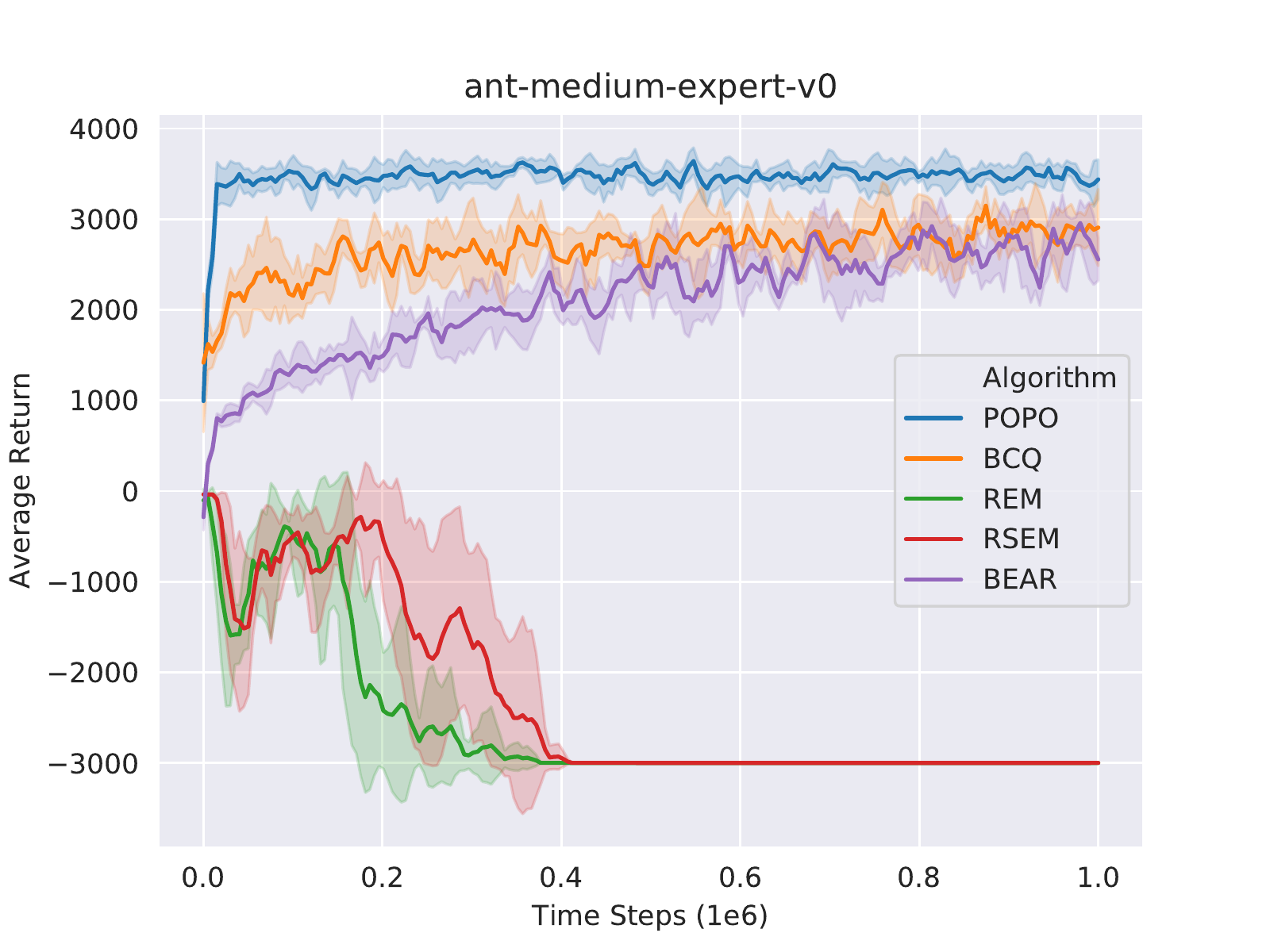}
	}
	\hspace{-0.3in}
	\subfigure{
		\includegraphics[width=1.7in]{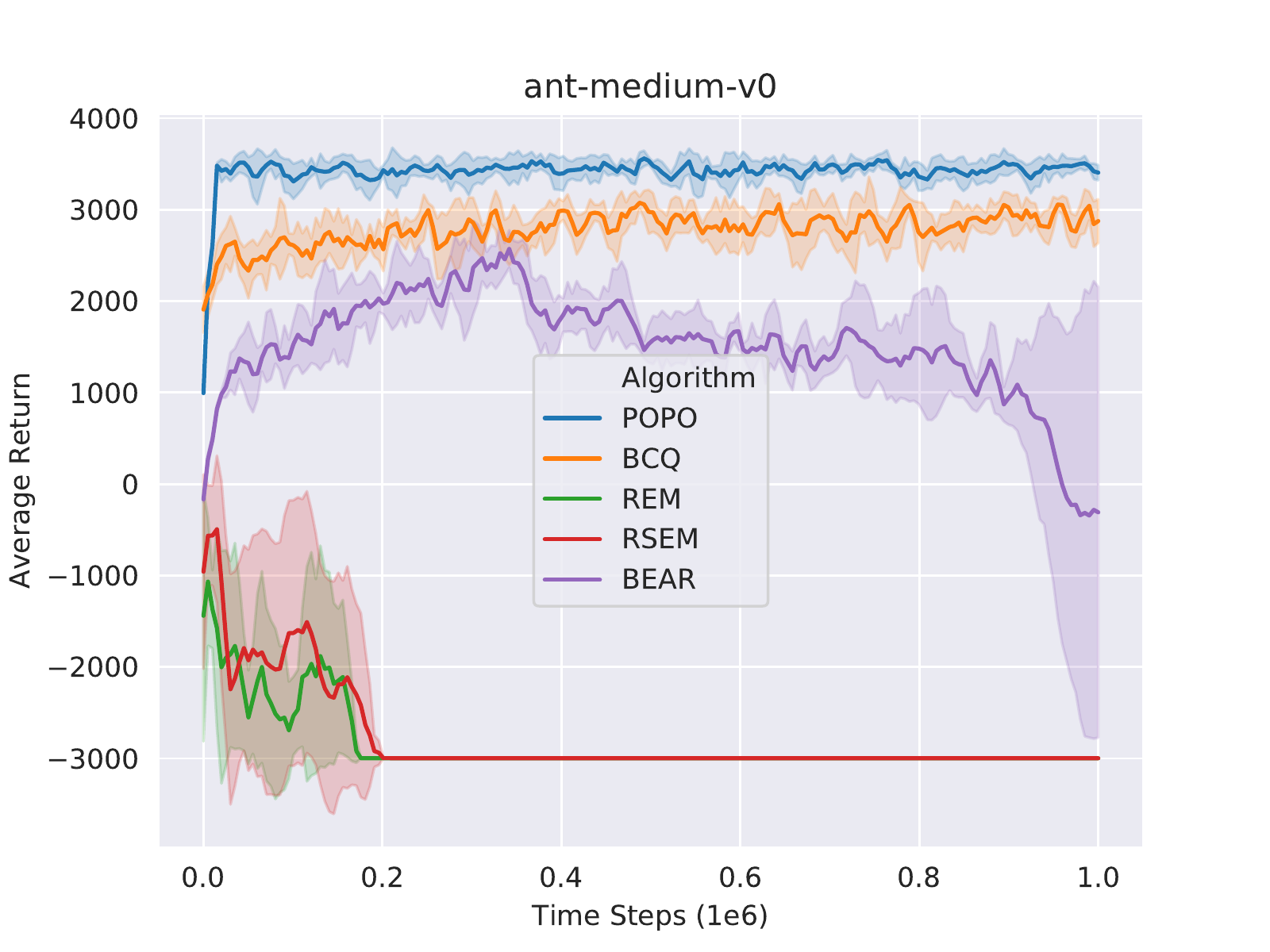}
	}
	\caption{Performance curves for OpenAI gym continuous control tasks in the MuJoCo suite. The shaded region represents a standard deviation of the average evaluation over five seeds. The BCQ is stable when tested, but it is not as good as the POPO. BEAR suffers from performance decrease when training too much time. REM almost failed during testing.}
	\label{fig:POPO_performance}
\end{figure}
Now we introduce pessimistic value function estimation from distributional RL view. The distributional Bellman optimality operator is defined by:
\begin{equation}
\mathcal{T}Z^{\pi}(s,a) := r + \gamma Z^{\pi}(s', \argmax_{a'\in \mathcal{A}} \mathbb{E}Z(s',a')),
\label{eq:distributional bellman eq}
\end{equation}
where $s' \sim p(\cdot|s,a)$ and random return $Z^{\pi}(s,a) := \sum_{t=0}^{\infty}r(s_t,a_t)$, $s_0=s, a_0=a, s_t \sim p(\cdot|s_{t-1},a_{t-1}), a_t\sim \pi(\cdot|s_t)$, and $X=Y$ denotes that random variable $X$ and $Y$ have equal distribution.
Let $F^{-1}_Z(\tau)$ denotes the quantile function at $\tau \in [0,1]$ for random variable $Z$. We write $Z_{\tau}:=F^{-1}_Z(\tau)$ for simplicity. Similar to~\cite{iqn}, we model the state-action quantile function as a mapping from state-action to samples from certain distribution, such as $\tau \sim U(0,1)$ to $Z_\tau(s,a)$. Let $\beta: [0,1] \rightarrow [0,1]$ be a distortion risk measure. Then the distorted expectation of random variable $Z(s,a)$ induced by $\beta$ is: 
\begin{equation}
Q_{\beta}(s,a;\theta) := \mathbb{E}_{\tau \sim U(0,1)}[Z_{\beta(\tau)}(s,a;\theta)].
\label{eq:Q_beta}
\end{equation}
We also call $Z_{\beta}$ critic. By choosing different $\beta$, we can obtain various distorted expectation, i.e., different attitude towards the estimation value. To avoid the abuse of symbols, $\tau$ in the following marks $\tau$ acted by $\beta$.
For the critic loss function, given two samples, $\tau, \tau' \sim U(0,1)$, the temporal difference error at time step $t$ is:
\begin{equation}
\Delta_t^{\tau, \tau'} = r_t+\gamma Z_{\tau'}(s_{t+1}, \pi_{\beta}(s_{t+1})) - Z_{\tau}(s_t, a_t).
\label{eq:distributional TD}
\end{equation}
Then the critic loss function of POPO is given by:
\begin{equation}
\mathcal{L}(s_t,a_t,r_t,s_{t+1}) = \frac{1}{N'}\sum_{i=1}^{N}\sum_{j=1}^{N'}\rho^{\kappa}_{\tau_i}(\Delta_t^{\tau_i,\tau'_{j}}),
\label{eq:Z loss}
\end{equation}
where
\begin{equation}
\begin{aligned}
\rho^{\kappa}_{\tau}(x) = |\tau - \mathbb{I}\{x<0\}| \frac{\mathcal{L}_{\kappa}(x)}{\kappa}, \quad \text{with}\\
\end{aligned}
\end{equation}
\begin{equation*}
\mathcal{L}_{\kappa}(x) = \left\{
\begin{aligned}
&\frac{1}{2}x^2,  \quad  & \text{if} |x|\leq \kappa \\
&\kappa(|x| - \frac{1}{2}\kappa), \quad & \text{otherwise} \\
\end{aligned}
\right.
\end{equation*}
in which $N$ and $N'$ is the number of i.i.d. samples $\tau_i, \tau'_j$ draw from $U(0,1)$ respectively.
Thus, given Z function, we can recover the $Q_\beta(s,a)$ from the Equation~\ref{eq:Q_beta}, further guides the learning process of policy.
\subsection{Distribution-Constrained Optimization }
To tackle OOD actions, we introduce a generative model, specifically, conditional Variational Auto-Encoder (VAE) $G(\cdot;\omega)$, consists of Encoder $E(\cdot|\cdot;\omega_1)$ and Decoder $D(\cdot|\cdot;\omega_2)$. Furthermore, VAE could constrain the distance between the actions sampled from the learned policy and that provided by the datasets. VAE reconstructs action on condition state $s$. We call the action produced by the VAE the central action $\hat{a}$. Thus, the loss function of VAE is:
\begin{equation}
	\begin{aligned}
	\mathcal{L}_{\text{VAE}} = \mathbb{E}_{s,a}\big[(a-\hat{a})^2+\frac{1}{2}D_{\text{KL}}(\mathcal{N}(\mu,\Sigma)\|\mathcal{N}(0,I)\big].
	\end{aligned}
\end{equation}
where $s,a \sim \mathcal{D}$ and $\hat{a}=D(z|s;\omega_2)$. To generate actions $a'$ w.r.t. state $s'$, firstly we copy the action $n$ times and send it to VAE in order to incorporate with policy improvement. Then we feed the actor network with central action $\hat{a'_i}=D(z_i|s';\omega_2)$ and state $s'$, then the actor network $\pi(\cdots;\phi)$ outputs a new action $\bar{a'_i}$. Combining $\hat{a'_i}$ and $\bar{a'_i}$ with residual style with coefficient $\xi$, we get the selected action $\tilde{a'_i}$. We choose action of $n$ output with highest value as the final output:
\begin{equation}
a'_{\text{new}} = \argmax_{a_i} Q_{\beta}(s',\tilde{a'_i};\theta),
\label{eq:a_new}
\end{equation}
where $\{\tilde{a'_i} = (\pi \circ D)(z_i|s')\}_{i=1}^{n}$.
We call this action generation method the residual action generation. The whole model structure is shown in Figure~\ref{fig:dataflow}. We use the DPG method (Equation~\ref{eq:DDPG_Value_function_update}) to train actor network $\pi$.
The benefits of residual action generation are apparent. In this way, for a given state, the generated action can be close to the actions contained in the data set with a similar state. At the same time, residual action generation maintains a large potential for policy improvement. 
We summarize the entire algorithm on Algorithm~\ref{algo:POPO}.
\begin{algorithm}[ht]
	\caption{Pessimistic Offline Policy Optimization (POPO)}
	\label{algo:POPO}
	\begin{algorithmic}
		\STATE \textbf{Require}
		\begin{itemize}
			\item Data set $\mathcal{D}$, num of quantiles $N$, target network update rate $\eta$, coefficient $\xi$.
			\item Distortion risk measure $\beta$, random initialized networks and corresponding target networks, parameterized by $\theta'_i \leftarrow \theta_i, \phi' \leftarrow \phi$, VAE $G=\{E(\cdot,\cdot;\omega_1),D(\cdot,\cdot;\omega_2)\}$. 
		\end{itemize}
		\FOR{iteration $= 1,2,...$}
		\STATE Sample mini-batch data $(s,a,r,s')$ from data set $\mathcal{D}$.
		\STATE $\#$ Update VAE
		\STATE $\mu, \sigma=E(a|s;\omega_1)$, $\hat{a}=D(z|s,;\omega_2)$, $z \sim \mathcal{N}(\mu,\sigma)$
		\STATE $\omega \leftarrow \argmin_{\omega}\sum(a-\hat{a})^2+\frac{1}{2}D_{\text{KL}}(\mathcal{N}(\mu,\Sigma)\|\mathcal{N}(0,I)$.
		\STATE $\#$ Update Critic.
		\STATE Set Critic loss $\mathcal{L}(\cdots;\theta)$ (Equation \ref{eq:Z loss}).
		\STATE $\theta \leftarrow \argmin_{\theta} \mathcal{L}(\cdots;\theta)$.
		\STATE $\#$ Update actor
		\STATE Generate $a_{\text{new}}$ from Equation~\ref{eq:a_new}
		\STATE $\phi \leftarrow \argmax_{\phi} Q_{\beta} (s, a_{\text{new}})$
		\STATE $\#$ Update target networks
		\STATE $\theta'_i \leftarrow \eta \theta_i + (1-\eta) \theta'_i$, $\phi'\leftarrow \eta\phi + (1-\eta)\phi'$  
		\ENDFOR
	\end{algorithmic}
\end{algorithm}
\section{EXPERIMENTS}
\begin{figure}[h]
	\centering
	\subfigure{
		\includegraphics[width=1.7in]{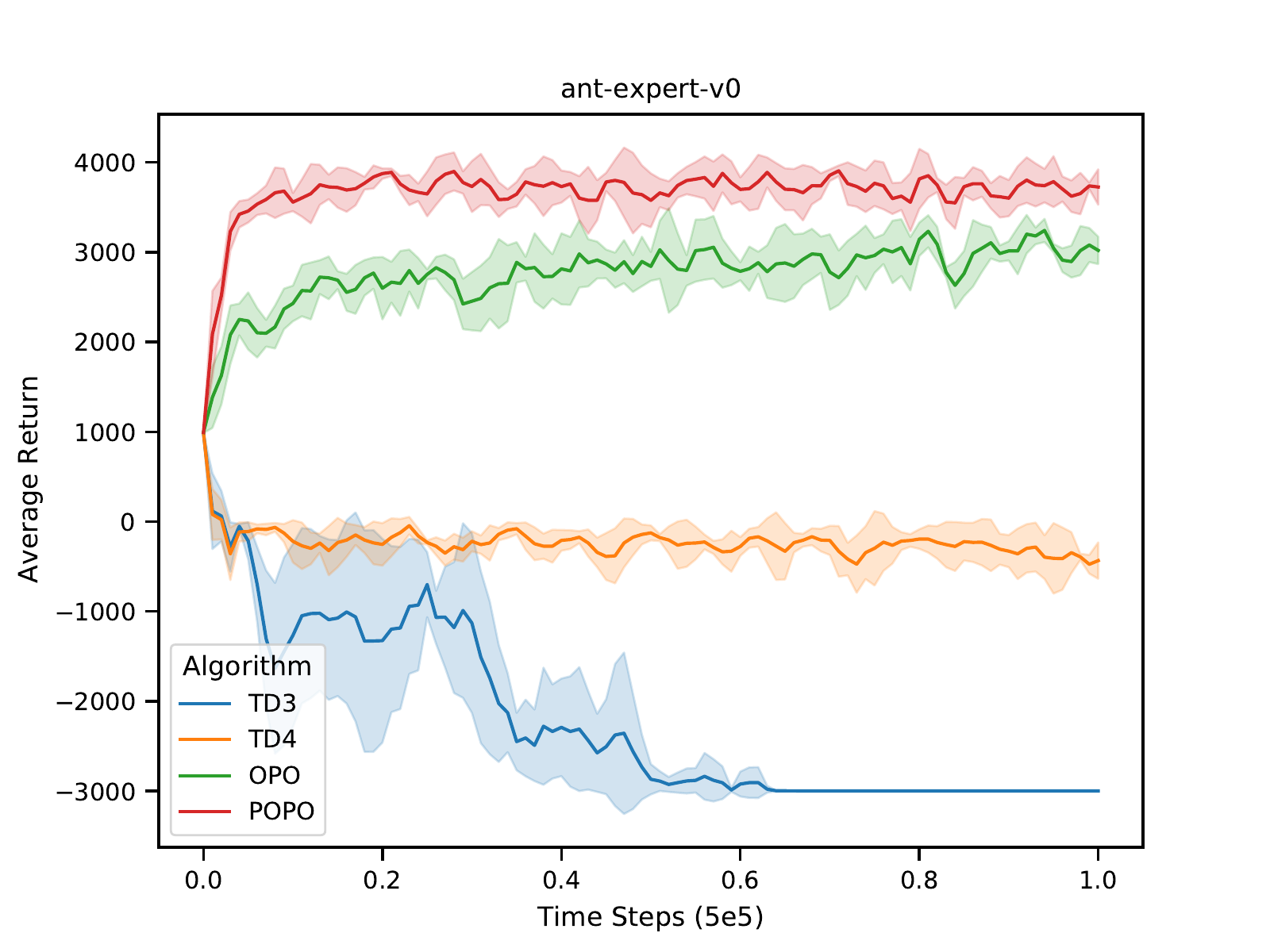}
	}  
	\hspace{-0.3in}
	\subfigure{
		\includegraphics[width=1.7in]{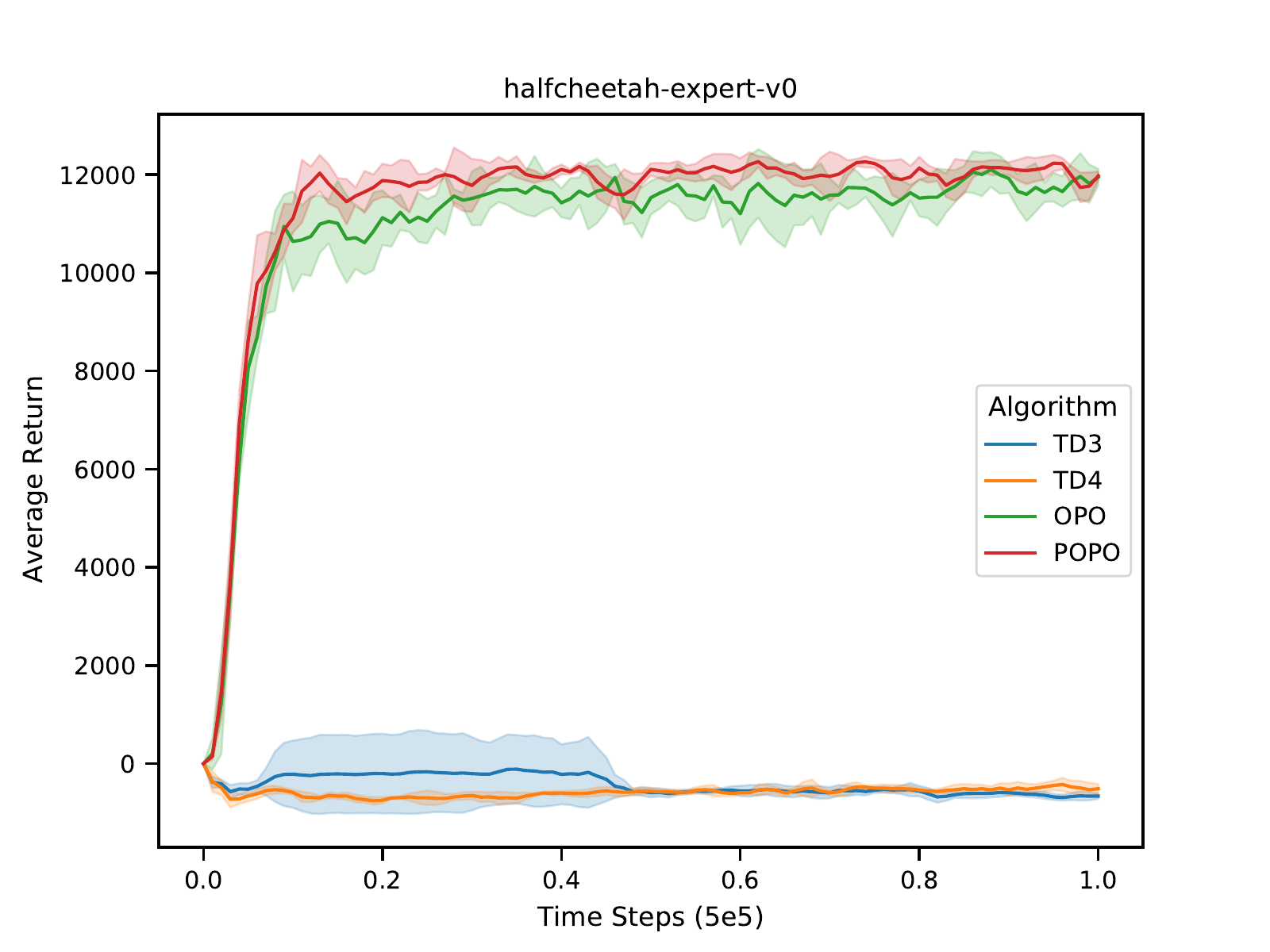}
	}
	\caption{Performance curves for ablation study. The shaded region represents a standard deviation of the average evaluation. The results show that the pessimistic critic does have improvement than the original TD3 algorithm. VAE makes offline optimization successful because it deals with the OOD action issue. The combination between VAE and pessimistic critic would produce better results.}
	\label{fig:ablation_vae_distributional}
\end{figure}

To evaluate our algorithm, we measure the performance of POPO on OpenAI gym~\cite{gym} continuous control tasks. We utilize the various quality original transitions from the d4rl datasets~\cite{d4rl} to train our model. "Expert" means the dataset is generated by a fine-tuned RL policy. "Medium-expert" marks the dataset is produced by mixing equal amounts of expert demonstrations and suboptimal data. The "medium" dataset consists of data generated by the suboptimal policy. Given the recent concerns about algorithms reflect the principles that informed its development~\cite{implementation}~\cite{drl_matters}, we implement POPO without any engineering tricks so that POPO works as we originally intended for. We compare our algorithm with the recently proposed SOTA offline RL algorithms BCQ\footnote{\url{https://github.com/sfujim/BCQ}.}, REM\footnote{\url{https://github.com/agarwl/off_policy_mujoco}}, and BEAR\footnote{\url{https://github.com/rail-berkeley/d4rl_evaluations}}. We use the authors' official implementations. The performance curves are graphed in Figure~\ref{fig:POPO_performance}. Note that we have not performed fine-tuned to the POPO algorithm due to the limitation of computing resources. Nevertheless, the results show that POPO matches or outperforms all compared algorithms.
\begin{table}[htbp]
	\caption{Distortion Risk Measures}
	\renewcommand\tabcolsep{3.0pt} 
	\begin{center}
		\begin{tabular}{ccc}
			\toprule
			\textbf{Wang($\zeta, \tau$)} & \textbf{CPW($\zeta, \tau$)}&\textbf{CVaR($\zeta, \tau$)}\\
			\midrule
			$\varphi(\varphi^{-1}(\tau)+\zeta)$ & $\frac{\tau^{\zeta}}{(\tau^{\zeta}+(1-\tau)^{\zeta})^{\frac{1}{\zeta}}}$ & $\zeta \tau$\\
			\bottomrule 
		\end{tabular}
		\label{table:distortion risk measure}
	\end{center}
\end{table} 
\subsection{Ablation Study}
The main modifications of POPO are VAE and pessimistic distributional critic. To explore the rule of each component, we design the ablation study. We call the POPO version without pessimistic distributional value function OPO, meaning OPO adopts TD3 style value function. We call the POPO version without VAE TD4 since it is TD3  with a pessimistic critic. Also, we use the original TD3 algorithm as a baseline because it neither has VAE nor a pessimistic critic. In this experiment, all the algorithms are tested over four seeds. The performance curves are graphed in Figure~\ref{fig:ablation_vae_distributional}. The results show that the pessimistic distributional value function does have a robust performance improvement compared to TD3 and OPO. Besides, VAE makes the offline RL successful because it solves the OOD actions issue. The combination between VAE and pessimistic critic would produce better results. Because VAE brings significant performance improvements, the pessimistic value function's positive impact on performance is relatively small but still significant.
\begin{figure}[h]
	\centering
	\subfigure{
		\includegraphics[width=1.7in]{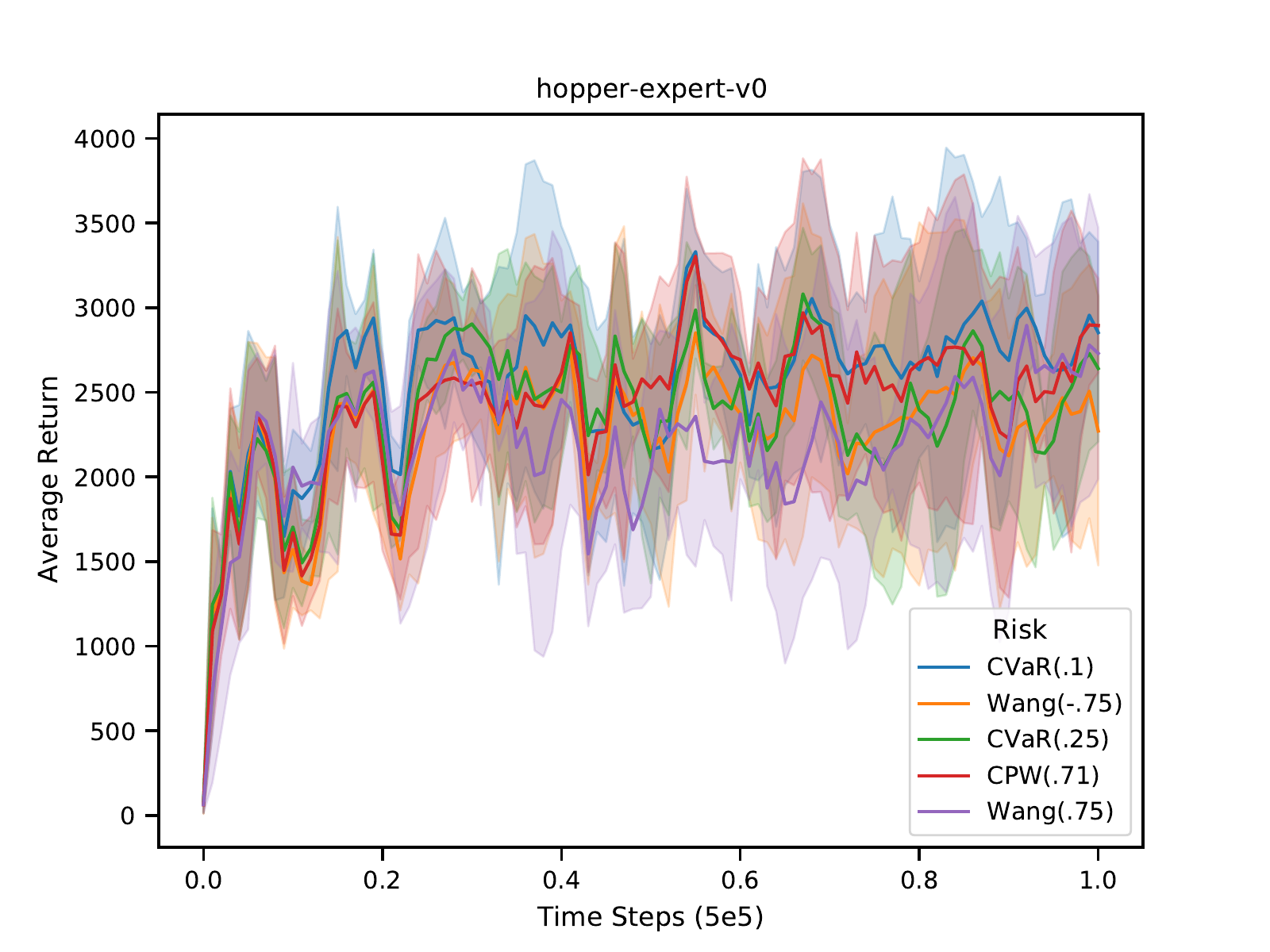}
	}  
	\hspace{-0.3in}
	\subfigure{
		\includegraphics[width=1.7in]{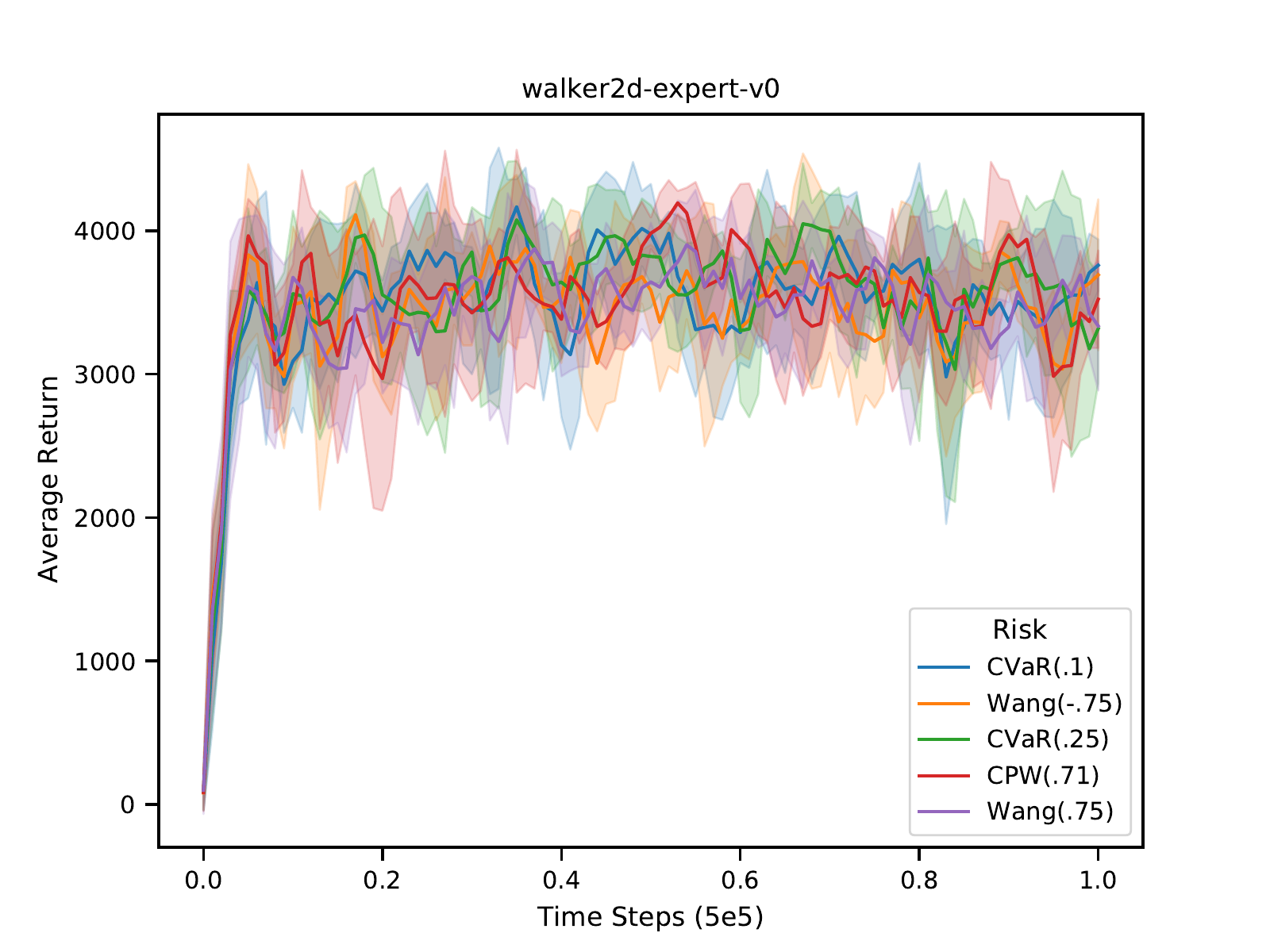}
	}
	\caption{Performance curves for various distortion risk measures. The result shows POPO with a pessimistic critic does perform better than one with an optimistic one.} 
	\label{fig:Ablation_variouts_distributional_function}
\end{figure}
\subsection{Pessimistic Value Function}
We study various distortion risk measure's effects on POPO. We select five different measures, CVaR$(0.1)$, Wang$(-0.75)$, CVaR$(0.25)$, CPW$(0,71)$ and Wang$(0.75)$, which rank pessimism from high to low~\cite{iqn}. Note that in $\text{Wang}(\zeta,\tau)$, $\varphi$ and $\varphi^{-1}$ is the standard Gaussian CDF and its reverse respectively. This measure produces an optimistic estimation for $\zeta>0$. We summarize various distortion risk measures in Tabel~\ref{table:distortion risk measure}. The performance curves are graphed in Figure~\ref{fig:Ablation_variouts_distributional_function}, which shows that POPO with pessimistic critic does perform better than one with an optimistic critic.
\subsection{Implementation Details}
\begin{table}[htbp]
	\caption{Hyper-parameters Setting}
	\renewcommand\tabcolsep{3.0pt} 
	\begin{center}
		\begin{tabular}{ccccccccc}
			\toprule
			\textbf{$\gamma$}&\textbf{$\xi$}&$\eta$&$\zeta$&n&batch-size&$\text{LR}_{\text{VAE}}$&$\text{LR}_{\text{Critic}}$&$\text{LR}_{\text{Actor}}$\\
			\midrule
			0.99 & 0.05 & $5\text{e}-3$ & $-0.75$&10 & 256& $3\text{e}-4$& $3\text{e}-4$& $3\text{e}-4$\\
			\bottomrule 
		\end{tabular}
		\label{table:hyper-paramers}
	\end{center}
\end{table}
When evaluating the average return, we freeze the whole model. Besides, to achieve a fair comparison, we evaluate all the algorithms on the default reward functions and environment settings without any changes. We implement POPO in the simplest form, without any engineering skills to gain performance. Unless specifically explained, all experiments are tested on four arbitrary random seeds. For distortion risk measure $\beta$, we use a pessimistic method $\text{Wang}(-0.75)$~\cite{wang2018exponentially}. We use a two-layer feedforward neural network for the actor and critic of POPO; each layer has 256 units, using the rectified linear units (ReLU) as the activate function for all hidden layers. A tanh activation function follows the last layer of the actor. The actor's output is the last layer's output multiplied by a hyper-parameter $ \xi $ and plus the input action and then clipped by the max action of environment. For the VAE, we use two-layer feedforward neural network for both Encoder and Decoder; each hidden layer has 750 units, using ReLU as activation. The dimension of the VAE latent variable is the 2$\times$ action space dimension. We use 32 quantiles for the critic. We list the other parameters in Table~\ref{table:hyper-paramers}. 
We evaluate performance every $5,000$ timesteps. The tested agent cannot collect the tested data or updates its parameters. 
\section{CONCLUSIONS}
In this work, we study why off-policy RL methods fail to learn in offline setting and propose a new offline RL algorithm. Firstly, we show that the inability to interact with the environment makes offline RL unable to eliminate the estimation gap through the Bellman equation. We conduct fine-grained experiments to verify the correctness of our theory. Secondly, We propose the Pessimistic Offline Policy Optimization (POPO) algorithm, which learns a pessimistic value function to get a strong policy. Finally, we demonstrate the effectiveness of POPO by comparing it with SOTA offline RL methods on the MuJoCo locomotion benchmarks datasets. 
\newpage
\bibliographystyle{./IEEEtran}
\bibliography{./IEEEabrv,root}

\end{document}